\documentclass[10pt,twocolumn,letterpaper]{article}

\usepackage{iccv}
\usepackage{times}
\usepackage{epsfig}
\usepackage{graphicx}
\usepackage{amsmath}
\usepackage{amssymb}
\usepackage[ruled,vlined]{algorithm2e}
\usepackage{algorithmic}
\usepackage{bm}
\usepackage{array}
\usepackage{mathtools}
\usepackage{booktabs}
\usepackage{graphicx}
\usepackage{color}
\renewcommand{\algorithmicrequire}{ \textbf{Input:}} %Use Input in the format of Algorithm
\renewcommand{\algorithmicensure}{ \textbf{Output:}} %Use Output in the format of Algorithm
\DeclareMathOperator{\sign}{sgn}

\newcommand{\xb}{\mathbf{x}}

\newcommand{\wb}{\mathbf{w}}

\newcommand{\argmin}{\operatornamewithlimits{argmin}}

\newcommand\blfootnote[1]{%
  \begingroup
  \renewcommand\thefootnote{}\footnote{#1}%
  \addtocounter{footnote}{-1}%
  \endgroup
}

\usepackage[pagebackref=true,breaklinks=true,letterpaper=true,colorlinks,bookmarks=false]{hyperref}

\iccvfinalcopy % *** Uncomment this line for the final submission

\def\iccvPaperID{1427} % *** Enter the ICCV Paper ID here

\ificcvfinal\fi
\begin{document}

%%%%%%%%% TITLE
\title{Relaxed Multiple-Instance SVM with Application to Object Discovery}
\author{Xinggang Wang$^\dag$, Zhuotun Zhu$^\dag$, Cong Yao, Xiang Bai$^*$\\
School of Electronic Information and Communications\\
Huazhong University of Science and Technology\\
{\tt\small xgwang@hust.edu.cn, zhuzhuotun@hust.edu.cn, yaocong2010@gmail.com, xbai@hust.edu.cn}
}

\maketitle
%\thispagestyle{empty}

%%%%%%%%% ABSTRACT

%%%%%%%%% BODY TEXT
\begin{abstract}

Multiple-instance learning (MIL) has served as an important tool for a wide range of vision applications, for instance, image classification, object detection, and visual tracking.
In this paper, we propose a novel method to solve the classical MIL problem, named relaxed multiple-instance SVM (RMI-SVM). We treat the positiveness of instance as a continuous variable, use Noisy-OR model to enforce the MIL constraints, and jointly optimize the bag label and instance label in a unified framework.
The optimization problem can be efficiently solved using stochastic gradient decent.
The extensive experiments demonstrate that RMI-SVM consistently achieves superior performance on various benchmarks for MIL. Moreover, we simply applied RMI-SVM to a challenging vision task, common object discovery. The state-of-the-art results of object discovery on Pascal VOC datasets further confirm the advantages of the proposed method.

\end{abstract}

%%%%%%%%% BODY TEXT
\section{Introduction}

Exploring big visual data is a new trend in computer vision in recent years \cite{TangCVPR14,MfMIL,chen2014enriching}. Especially, with the development of deep learning, the performances of many large-scale visual recognition tasks have been significantly improved. However, the supervised deep learning methods, \emph{e.g.}, deep convolutional neural networks (DCNN) \cite{krizhevsky2012imagenet}, rely heavily on the huge number of human-annotated data that are non-trivial to get. Finely labeled images/videos, which have pixel-level labels and bounding-box labels, are very limited and expensive. However, there are hundreds times of weakly labeled visual data that have image-level labels or noisy labels. For example, we can extract image label from its text caption on Flickr \cite{Fang2015}. How to use the weakly labeled visual data for object recognition is a quite important research problem.

\blfootnote{$^\dag$ equal contribution; $^*$ corresponding author.}

\begin{figure}
  \centering
  \includegraphics[width=0.5\textwidth]{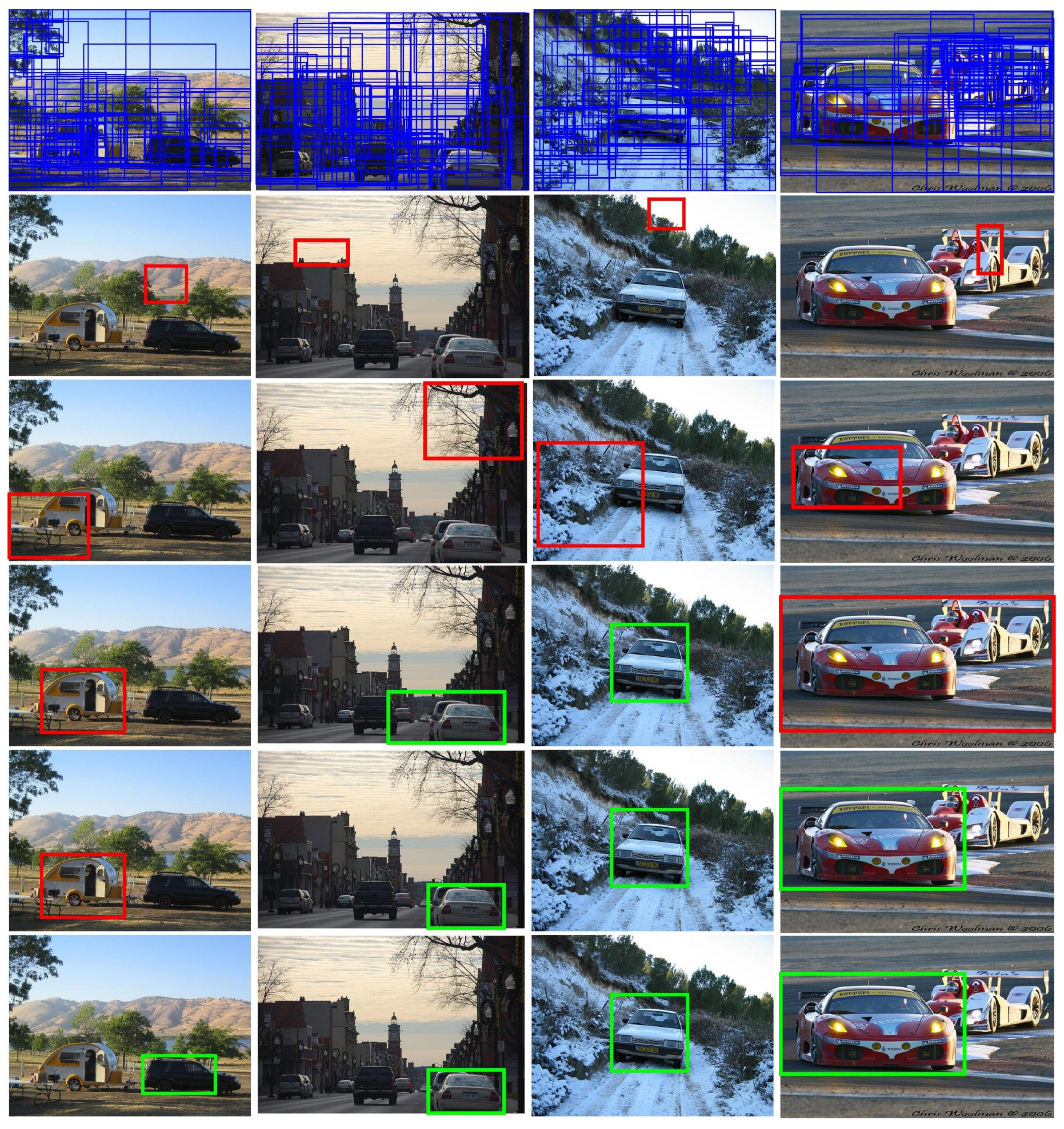}\\
  \caption{Iteratively discover the locations of objects using the proposed RMI-SVM algorithm. 1st row: The top 100 object proposals detected by Edgebox \cite{Edgebox}. 2nd row: Randomly initialized object locations in iteration 0. 3rd - 6th rows: The detected object locations in iteration 100, 500, 1000, and 2000, respectively. The blue boxes show the object proposals, the red boxes show the detected objects that do not enough overlap with ground-truth, and the green boxes show the detected objects that own enough overlap with ground-truth. (Best viewed in color.)}\label{fig:intro}
  \vspace{-0.7cm}
\end{figure}

The multiple-instance learning (MIL), proposed by Dietterich \emph{et al.}~\cite{DietterichLL97} for the purpose of drug activity prediction, is a popular tool for exploring sematic information in weakly labeled visual data. In MIL, instead of being given the labels of each individual instance, the learner receives a set of labeled bags, each containing plenty of instances. In the binary-classification task, a bag may be labeled as positive if \emph{at least} one instance is positive. On the other hand, a bag will be labeled as negative if \emph{none} of the instances is positive. Typically, we can regard an image/video as a bag, and a patch/cube inside as an instance. Objects of interest are considered as positive instances, and the rest are considered as negative instances. Besides of learning bag distribution, we expect MIL can infer the label of instance to find objects of interest. However, not all MIL algorithms can reach this goal; most of them only focus on bag classification \cite{DietterichLL97,migraph,miFV}.

Selecting positive instances and learning a discriminative/generative instance model to classify bag is a popular way for solving MIL problem in computer vision. For example, online multiple-instance Boosting was applied for robust visual tracking in \cite{Babenko2011}; multiple instance SVM \cite{andrews2002support} was used to learn deformable object detector \cite{felzenszwalb2010object}, which is also called latent SVM; and, unsupervised multiple instance Boosting was developed for multi-class learning in \cite{bMCL}. However, these existing methods all treat instance selection and model learning as two separated procedures, and use EM-style algorithm for optimization.
In this paper, we propose a unified framework to jointly optimize the label of instance and learn instance model by taking the advantage of relaxing the discrete instance label and stochastic gradient descent. The MIL constraints are formulated using a Noisy-OR model. The instance model is a simple linear SVM model which allows fast training and prediction. The optimization problem can be efficiently solved using stochastic gradient descend algorithm, and is very robust to initialization in practical applications.

As shown in Fig.~\ref{fig:intro}, the proposed MIL algorithm can be applied to object discovery, which is also called weakly-supervised object location and object co-localization. At first, we obtain hundreds of object proposals using the Edgebox \cite{Edgebox} and extract the deep feature for each proposal using DCNN \cite{krizhevsky2012imagenet} in each image. Then, The proposed RMI-SVM algorithm is able to gradually find the true object location from the initialization location which is randomly selected. In the procedure of training RMI-MIL, we get exact object locations; besides, the learned instance model (object model) can be even used for object detection in unseen images. Our object discovery method is clean, simple but effective. It uses the off-shelf Edgebox object proposals and DCNN features. After feature extraction is done, it takes about 35 minutes using a single CPU to discover all the 20 classes in the Pascal VOC 2007 dataset. In the experiments, RMI-SVM shows superior performance when compared to both other MIL algorithms and the state-of-the-art object discovery methods.

To summarize, our main contributions are three folds: 1) a novel MIL formulation that relaxes the MIL constraints into convex program; 2) a fast and robust MIL solution via SGD; 3) an effective weakly-supervised object discovery based on the proposed RMI-SVM, which can obtain the state-of-the-art performance on the challenging Pascal 2007 dataset.

\section{Related Work}

Multiple-instance learning was firstly proposed by Dietterich \emph{et al.}~\cite{DietterichLL97} for drug activity prediction. After that, since it is very useful in both machine learning and computer vision, lots of MIL algorithms have been proposed. Some of the typical methods are briefly introduced as follows:
The diverse density (DD) method \cite{maron1998framework} tackles MIL by finding regions in the instance space with instances from many different positive bags and few instances from negative bags. In \cite{EMDD}, DD is refined using expectation maximization (EM). In DD-SVM~\cite{ChenW04}, instance prototype is extracted based on DD function in the instance feature space, followed by a nonlinear mapping to project each bag to a point in the bag feature space. miSVM and MILBoost were proposed in \cite{andrews2002support} and \cite{zhang2005multiple} in which they train SVM and boosting classifier for instances respectively.
Recent work on MIL includes: representing the bags as graphs and explicitly modeling the relationships between instances within a bag in \cite{migraph}, studying the problem if there are infinite number of instances in a bag in \cite{babenko2011multiple}, mining key instances from a citer kNN graph for bag classification \cite{keyinstance}, building a deep learning framework in a weakly supervised setting~\cite{wu2015deep}, and using bag-of-word model to solve large-scale MIL problem \cite{miFV}.

MIL is highly related to and plays an important role in many visual recognition tasks, especially in weakly-supervised object discovery, for example, person head discovery \cite{zhang2005multiple}, object part discovery \cite{DollarECCV08mcl,felzenszwalb2010object}, object class discovery \cite{bMCL}.
For generic object discovery in the wild, MIL also works very well. A generative and convex MIL algorithm was proposed in \cite{RobustSubspace} for object discovery based salient object detection.
Very recently, MIL is trained on the top of DCNN to discover object for automatically image captioning \cite{Fang2015}.

Object discovery has recently drawn lots of attentions. Top-down segmentation priors based object detector is combined for pixel-level object discovery in \cite{chen2014enriching}. A part-based matching between object proposals is proposed for unsupervised object discovery in \cite{cho2015unsupervised}. A multi-fold MIL is designed for object discovery in \cite{MfMIL}. And, a joint box-image formulation is proposed in \cite{TangCVPR14} and applied for large-scale object discovery on the ImageNet dataset. Different from the existing object discovery methods, our object discovery method utilizes the proposed novel RMI-SVM, Edgebox and off-the-shelf DCNN feature to construct an end-to-end system, in which all the components are very efficient and effective.

\section{Relaxed Multiple-Instance SVM}\label{RMI-SVM}

% The proposed approach takes a set of ambiguous examples $\mathcal{D}=\lbrace (\bm{X_i},Y_i), i=1,2,..., n\rbrace$ as input. Each \emph{labeled} bag $\bm{X_i}$ consists of a set of \emph{unlabeled} instances $x_{ij}$, that is to say $ \bm{X_i}=\lbrace x_{i1},{x_{i2}},..., x_{im_i}\rbrace $, where $m_i$ denotes the number of instances in the bag $\bm{X_i}$.

\subsection{MIL Relaxations}

We first give notation of MIL as preliminaries. In MIL, we are given a set bags $X = \{ X_1, \dots, X_n \}$; each bag is consisted with a set of instance $X_i = \{ \xb_{i1}, \dots, \xb_{im_i} \}$, where $m_i$ denotes the number of instances in the bag $X_i$; and each instance is represented by a $d$-dimensional vector $\xb_{ij} \in \mathbf{R}^{d \times 1}$.
Each bag is associated with a bag label $Y_i \in \{ 0, 1\}$; and each instance is associated with an instance label $y_{ij} \in \{ 0, 1 \}$ too.
The relation between bag label and instance labels, which is also called \textit{MIL constraints},  is interpreted in the following way:
\begin{itemize}
  \item If $Y_i = 0$, then $y_{ij} = 0$ for all $j \in [1, \dots, m_i]$, \ie, no instance in the bag is positive.
  \item If on the hand $Y_i = 1$, then at least one instance $\xb_{ij} \in X_i$ is a positive instance of the underlying concept.
\end{itemize}

In RMI-SVM, we relax the instance label $y_i$ to be a continues variable in the range of $[0,1]$, which is the probability of $\xb_{ij}$ being positive, denoted as $p_{ij}$. Without loss of generality, we use a linear model as instance model. $p_{ij}$ is given by a logistic function
\begin{equation}\label{eq:pij}
   p_{ij}=Pr(y_{ij}=1|\xb_{ij};\wb)=\frac{1}{1+e^{-\wb^T\xb_{ij}}},
\end{equation}
where $\wb$ is the weight vector of the linear model which needs to be optimized through in our formulation.

Only knowing the positive probability of instances is far from enough since the final goal of MIL is to predict whether a bag is positive. And we only know the bag-level label, but do not know the instance-level label. To bridge the gap between instance level and bag level, we adopt the Noisy-OR(NOR) model. The probability of bag regarded as positive is computed via
\begin{equation}\label{eq:NOR}
    P_i=Pr(Y_{i}=1|{X_i};\wb)=1-\prod_{j=1}^{m_i}(1-p_{ij}).
\end{equation}
Assuming that one instance in the bag is predicted as positive, \emph{e.g.}, $p_{ij}=1$, then we can find $P_i=1$ according to Eq.\eqref{eq:NOR}. If all the instances in the bag are predicted as zero, we can find $P_i=0$. The NOR model is a relaxed version of the MIL constraints.

\subsection{Objective Function}

The above relaxations make the MIL problem more tractable, because there is no discrete variable and all parts in Eq.~\eqref{eq:LossFunc} are differentiable.
Considering the instance-level loss, bag-level loss, and model regularization, we give our MIL objective function as follows:
\begin{equation}\label{eq:LossFunc}
\min_{\wb} \frac{\lambda}{2}{\lVert \wb \rVert}^2+\frac{\beta}{n}\sum_{i=1}^n\mathcal{L}_{bag_i}+\frac{1}{n}\sum_{i=1}^n\frac{1}{m_i}
\sum_{j=1}^{m_i} \mathcal{L}_{ins_{ij}}, \\
\end{equation}
where the first regularization item is to avoid overfitting; $\mathcal{L}_{bag_i}$ denotes the cost item for
$i$-$th$ bag prediction and $\mathcal{L}_{ins_{ij}}$ denotes the cost item for $ij$-$th$ instance prediction. More specifically, they are denoted as
%\begin{equation}
%    \mathcal{L}_{bag}=-\sum_i^n \lbrace Y_i\log P_i+(1-Y_i)\log(1-P_i) \rbrace,
%\end{equation}
%\begin{equation}
%    \mathcal{L}_{ins}=\sum_i^n\sum_j^m max(0,[m_0-sgn(p_{ij}-0.5)w^Tx_{ij}]),
%\end{equation}
\begin{align}
 &\mathcal{L}_{bag_i}=-\lbrace Y_i\log P_i+(1-Y_i)\log(1-P_i) \rbrace,\label{eq:Lbaglr}  \\
 &\mathcal{L}_{ins_{ij}}=\max(0,[m_0-\sign(p_{ij}-p_0)\wb^T\xb_{ij}]).\label{eq:Lins}
\end{align}
where $\sign$ is the sign function; $m_0$ is a crucial margin parameter used to separate the positive instances and negative instances distant from the hyper line in the feature space; $p_0$ is a threshold parameter to determine positive or instance.

The goal of RMI-SVM is to find an optimal instance model to determine the label of instances and bags. Thereby, the optimal instance model is given by:
\begin{equation}
\wb^* = \argmin_{\wb} \frac{\lambda}{2}{\lVert \wb \rVert}^2+\frac{\beta}{n}\sum_{i=1}^n\mathcal{L}_{bag_i}+\frac{1}{n}\sum_{i=1}^n\frac{1}{m_i}
\sum_{j=1}^{m_i} \mathcal{L}_{ins_{ij}}.
\end{equation}
The positiveness of instance is given by $p_{ij}=\frac{1}{1+e^{-\wb^{*T}\xb_{ij}}}$. If $ p_{ij} \geq p_0$, $y_{ij}=1$; otherwise, $y_{ij}=0$.

%
%The cost function in instance level $\mathcal{L}_{ins}$ is given as
%
%\begin{equation}
%    \mathcal{L}_{ins}=-\sum_i^n\sum_j^m \lbrace p_{ij}\log p_{ij}+(1-p_{ij})\log(1-p_{ij})\rbrace,
%\end{equation}
%
%The overall cost is $\mathcal{L}_{MIL}$ defined as the sum of cost in bag and instance level,
%
%\begin{equation}
%    \mathcal{L}_{MIL}=(1-\lambda)\mathcal{L}_{bag}+\lambda \mathcal{L}_{ins},
%\end{equation}
%where $\lambda\in [0,1]$ is the factor parameter adjusting the contribution of bag level and instance level to the overall cost.

\subsection{Derivations}

The above optimization problem in Eq.~\eqref{eq:LossFunc} can be solved using stochastic gradient descent. Therefore,
we derive the partial derivative of $\mathcal{L}_{bag_i}$ and $\mathcal{L}_{ins_{ij}}$ to the weight vector $\wb$.

Using the chain rule of calculus in Eq.~\eqref{eq:Lbaglr}, the partial derivative of $\mathcal{L}_{bag_i}$ with respect to $\wb$
is derived as
\begin{equation}\label{eq:Lbag}
 \begin{split}
    \frac{\partial \mathcal{L}_{bag_i}}{\partial \wb}
    &=\frac{\partial \mathcal{L}_{bag_i}}{\partial P_{i}}\cdot\sum_{j=1}^{m_i}\frac{{\partial P_{i}}}{{\partial p_{ij}}}
    \frac{\partial p_{ij}}{\partial \wb},\\
%    &=-\sum_i^n{\lbrace \frac {Y_{i}}{P_i} -\frac {(1-Y_i)}{1-P_i}\rbrace}\cdot\frac{\partial P_{i}}{\partial p_{ij}}\cdot \
%frac{\partial p_{ij}}{\partial w_{k}}  \\
%    &=-\sum_i^n\frac {Y_i-P_i}{P_i(1-P_i)}\cdot\prod_{k\ne j}^m(1-p_{ik})\cdot \frac{\partial p_{ij}}{\partial w_{k}}\\
%    %&=\sum_i^n\frac {Y_i-P_i}{P_i(1-P_i)}\cdot \frac{1-P_i}{1-p_{ij}}\cdot p_{ij}(1-p_{ij})x_{ij}^k\\
%    &=-\sum_i^n\frac {Y_i-P_i}{P_i(1-P_i)}\cdot \frac{1-P_i}{1-p_{ij}}\cdot \frac{\partial p_{ij}}{\partial w_{k}}\\
%    &=-\sum_i^n\frac {Y_i-P_i}{P_i(1-p_{ij})}\cdot \frac{\partial p_{ij}}{\partial w_{k}}
%%    %&=\sum_i^n\frac {(Y_i-P_i)p_{ij}}{P_i}\cdot x_{ij}^k
 \end{split}
\end{equation}
where $\frac{\partial \mathcal{L}_{bag_i}}{\partial P_{i}}$ and $\frac{{\partial P_{i}}}{{\partial p_{ij}}}$ is given by
\begin{align}
& \frac{\partial \mathcal{L}_{bag_i}}{\partial P_{i}}=-{\lbrace \frac {Y_{i}}{P_i} -\frac {(1-Y_i)}{1-P_i}\rbrace}=-
\frac {Y_i-P_i}{P_i(1-P_i)};\label{eq:Lbagpart1}\\
&  \frac{{\partial P_{i}}}{{\partial p_{ij}}}=\prod_{\mathclap{ k=1, k\ne j}}(1-p_{ik})=\frac{\prod_{k=1}^{m_i}(1-p_{ik})}{(1-p_{ij})}=
\frac{1-P_i}{1-p_{ij}}.\label{eq:Lbagpart2}
\end{align}
%\begin{equation}
% \begin{split}
% \frac{\partial \mathcal{L}_{bag_i}}{\partial P_{i}}
% &=-{\lbrace \frac {Y_{i}}{P_i} -\frac {(1-Y_i)}{1-P_i}\rbrace}=-\frac {Y_i-P_i}{P_i(1-P_i)}
% \end{split}
%\end{equation}
%According to Eq.~\eqref{eq:NOR},
%
%\begin{equation}
% \frac{{\partial P_{i}}}{{\partial p_{ij}}}=\prod_{k=1, k\ne j}^m(1-p_{ik})=\frac{\prod_j^m(1-p_{ij})}{(1-p_{ij})}=\frac{1-P_i}{1-p_{ij}}
%%A=\sum_{\mbox{\tiny$\begin{array}{c}
%%0\le i \le m\\
%%0\le j \le n\\
%%0\le k \le q\end{array}$}}a_{ijk}
%\end{equation}
%Therefore, the derivative of the cost function to weight of bag level is given by
According to Eq.~\eqref{eq:pij}, we can find the partial derivative of $p_{ij}$ to $w$ is
\begin{equation}\label{eq:pijpart}
 \begin{split}
  \frac{\partial p_{ij}}{\partial \wb}
  &=-(1+e^{-\wb^T\xb_{ij}})^{-2}\cdot e^{-\wb^T\xb_{ij}}\cdot (-\xb_{ij})\\
  %&={p_{ij}}^2\cdot({p_{ij}}^{-1}-1)\cdot \xb_{ij}\\
  &=p_{ij}(1-p_{ij})\cdot \xb_{ij}.
 \end{split}
\end{equation}
Appying Eq.(~\ref{eq:Lbagpart1},~\ref{eq:Lbagpart2},~\ref{eq:pijpart}) to Eq.~\eqref{eq:Lbag}, the final expressoin of partial
derivative of $\mathcal{L}_{bag_i}$ with respect to $\wb$ is
\begin{equation}
\frac{\partial \mathcal{L}_{bag_i}}{\partial \wb}=-\sum_{j=1}^{m_i}\frac {p_{ij}(Y_i-P_i)}{P_i}\xb_{ij}.
\end{equation}

As for the partial derivative of $\mathcal{L}_{ins_{ij}}$ with respect to $\wb$, this expression is derived as
\begin{equation}
 \frac{\partial \mathcal{L}_{ins_{ij}}}{\partial \wb}=-\mathbf{1}[\sign(p_{ij}-p_0)\wb^T\xb_{ij}<m_0]\cdot \sign(p_{ij}-p_0)\xb_{ij},
\end{equation}
where $\mathbf{1}[\sign(p_{ij}-p_0)\wb^T\xb_{ij}<m_0]$ is an indicator function which equals one if its argument is true and zero otherwise.

%%According to the Multiple Instance Learning, all instances in negative bags are labeled as zero.
%\begin{equation}
% \begin{split}
% \frac{\partial \mathcal{L}_{ins_{ij}}}{\partial w_{k}}
% &= \frac{\partial \mathcal{L}_{ins}}{\partial p_{ij}}\cdot\frac{{\partial p_{ij}}}{{\partial w_{k}}} \\
%% &=-\sum_i^n\sum_j^m\lbrace \log p_{ij} -\log (1-p_{ij})\rbrace \cdot \frac{\partial p_{ij}}{\partial w_{k}} \\
%% &=\sum_i^n\sum_j^m\log \frac{1-p_{ij}}{p_{ij}} \cdot \frac{\partial p_{ij}}{\partial w_{k}}
% \end{split}
%\end{equation}
%
%\begin{equation}
%  \frac{\partial \mathcal{L}_{ins}}{\partial p_{ij}}=-\sum_i^n\sum_j^m\lbrace \log p_{ij} -\log (1-p_{ij})\rbrace
%\end{equation}

%\begin{equation}
% \begin{split}
% \frac{\partial \mathcal{L}_{ins^+}}{\partial w_{k}}
% &= \frac{\partial \mathcal{L}_{ins}}{\partial p_{ij}}\cdot\frac{{\partial p_{ij}}}{{\partial w_{k}}} \\
% &=-\sum_i^n\sum_j^m\lbrace \log p_{ij} -\log (1-p_{ij})\rbrace \cdot \frac{\partial p_{ij}}{\partial w_{k}} \\
% &=\sum_i^n\sum_j^m\log \frac{1-p_{ij}}{p_{ij}} \cdot \frac{\partial p_{ij}}{\partial w_{k}}
% \end{split}
%\end{equation}

%Therefore, the updated procedure is given by%, {\color{red}wrong here}
%\begin{equation}
% \Delta {w_{k}}=\alpha\sum_i^n\sum_j^m\lbrace(1-\lambda) \frac {Y_i-P_i}{P_i(1-p_{ij})}-\lambda\log \frac{1-p_{ij}}{p_{ij}}
%\rbrace \cdot \frac{\partial p_{ij}}{\partial w_{k}},
%\end{equation}
%where the partial derivative of $p_{ij}$ to $w_{k}$ is

\subsection{SGD Optimization}

We describe the optimization method in this subsection and also provide the pseudo-code.
As mentioned in Sec.~\ref{RMI-SVM}, our method performs SGD on the objective in Eq.~\eqref{eq:LossFunc} with a varied learning rate strategy.
On a iteration $t$ in our algorithm, we randomly choose a bag $(X_{k_t},Y_{k_t})$ from the training sets $\mathcal{D}$ via picking an index $k_t\in \lbrace 1,2,...,n\rbrace$ in a standard uniform distribution. Then we change the objection in Eq.~\eqref{eq:LossFunc} to an approximation based on the sample bag, obtaining
\begin{equation}
f(\wb; X_{k_t})=\frac{\lambda}{2}{\lVert \wb \rVert}^2+\beta\mathcal{L}_{bag_{k_t}}+\frac{1}{m_{k_t}}\sum_{j=1}^{m_{k_t}}
\mathcal{L}_{ins_{k_tj}}.
\end{equation}
Considering the gradient of the approximate function, given by
\begin{align}
&\nabla_t =\frac {\partial f(\wb; X_{k_t})}{\partial \wb}=\lambda \wb-\sum_{j=1}^{m_{k_t}} \xb_{{k_t}j}\lbrace \beta\cdot\frac
{p_{{k_t}j}(Y_{k_t}-P_{k_t})}{P_{k_t}}+\nonumber \\
&\frac{\sign(p_{{k_t}j}-p_0)}{m_{k_t}}\cdot\mathbf{1}[\sign(p_{{k_t}j}-p_0)\wb^T\xb_{{k_t}j}<m_0]\rbrace,
\end{align}
we update the weight vector using a varied learning rate $\eta_t=1/[(t+1)\cdot \lambda]$, that is $\wb_{t+1}\gets \wb_t-\eta_t\cdot \nabla_t$. When $t$ reaches a predefined iteration $T$, we output the last weight $\wb_T$. It is worth noting that after each gradient update, we employ a projection operation of $\wb$ on the $L_2$ ball of radius $1/\sqrt{\lambda}$ just as mentioned in~\cite{Pegasos} via the following update,
\begin{equation}
\wb_{t+1}\gets \min \lbrace 1,\frac {1/\sqrt \lambda}{\lVert \wb_{t+1}\rVert}\rbrace \wb_{t+1}.
\end{equation}
This modification can significantly accelerate the rate of convergence in the optimization step.

In summary, the pseudo-code for solving RMI-SVM is given in Algorithm~\ref{alg:Framwork}, which is granted to get a local optimal solution for the objective function Eq.~\eqref{eq:LossFunc}. In practical application, it gives satisfactory accuracy and fast speed.

\begin{algorithm}[h]
\caption{Pseudo-code for solving RMI-SVM.\label{alg:Framwork}}
\DontPrintSemicolon
\KwIn {$\mathcal{D}$, $\lambda$, $\beta$, $p_0$, $m_0$, $T$\\}
\KwOut {$\wb_{T+1}$}
\Begin{
Initialize: Set $\wb_1=0$\;
\For{$t=1,2,...,T$}{
choose $k_t\in \lbrace 1,2,...,n\rbrace$, uniform distribution\\
Set $j^+=\lbrace j|\sign (p_{k_tj}-p_0)\wb_t\xb_{{k_t}j}<m_0\rbrace$\\
Set $m_{k_t}=\lvert X_{k_t}\rvert$\\
Set $\eta_t=\frac{1}{\lambda t}$\\
Set $\wb_{t+1}\gets\lbrace (1-\eta_t \lambda)\wb_t+\beta \eta_t \sum_j \xb_{k_tj}\frac {p_{{k_t}j}(Y_{k_t}-P_{k_t})}{P_{k_t}}+$\\
$\frac{\eta_t}{m_{k_t}} \sum_{j^+}\sign(p_{{k_t}j^+}-p_0)\xb_{k_tj^+}\rbrace$\\
Set $\wb_{t+1}\gets \min \lbrace 1,\frac {1/\sqrt \lambda}{\lVert \wb_{t+1}\rVert}\rbrace \wb_{t+1}$
{}
}
}
\end{algorithm}

\section{Experiments on MIL Benchmarks}

In this and the following section, we perform experiments to test RMI-MIL for bag classification on MIL benchmarks and object discovery in the wild, respectively.
RMI-MIL is implemented in MATLAB and experiments are carried out on a desktop machine with Intel(R) Core(TM) i7-3930K CPU (3.20GHz) and 64GB RAM. The code will be released on publication. In the following subsections, three widely-used MIL benchmarks on different applications are tested.

\subsection{Drug Activation Prediction}\label{sec:Drug}

The task is to predict whether a new drug molecule can bind well to a target protein, which is mainly determined by the shape of the molecule. A ``right" molecular shape can bind well to the target protein. Unfortunately, a molecule always exhibits multiple shapes. In this case, a good molecule will bind well if \emph{at least} one of its shapes is right, while a poor molecule will not bind well if \emph{none} of its shapes can bind. Therefore, the drug prediction task can be formulated as a MIL problem.

%\begin{table}[ht]
%\caption{Average prediction accuracy(\%) on benchmarks.}\label{table:MCdata}
%\begin{center}
%\begin{tabular}{lccccc}\toprule[1.5pt]
%%\hline
%Algorithm       &MUSK1          &MUSK2      &Elept   &Fox        &Tiger\\
%\hline
%RMI-SVM            &80.8           &82.4       &$\bm{87.8}$       &$\bm{63.6}$       &$\bm{87.9}$\\
%                &$\pm 2.0$      &$\pm1.6$   &$\bm{\pm0.7}$   &$\bm{\pm 2.8}$  &$\bm{\pm 0.9}$\\
%\hline
%MIGraph         &90.0           &90.0       &85.1       &61.2       &81.9\\
%                &$\pm 3.8$      &$\pm2.7$   &$\pm2.8$   &$\pm 1.7$  &$\pm 1.5$\\
%\hline
%miGraph         &88.9           &$\bm{90.3}$       &86.8       &61.6       &86.0\\
%                &$\pm 3.3$      &$\bm{\pm2.6}$   &$\pm0.7$   &$\pm 2.8$  &$\pm 1.6$\\
%\hline
%MI-Kernel       &88.0           &89.3       &84.3       &60.3       &84.2\\
%                &$\pm 3.1$      &$\pm1.5$   &$\pm1.6$   &$\pm 1.9$  &$\pm 1.0$\\
%\hline
%%MISVM           &77.9           &84.3       &81.4       &59.4       &84.0\\
%%miSVM           &87.4           &83.6       &82.0       &58.2       &78.9\\
%MISVM           &80.4           &-       &81.4       &57.8       &84.0\\
%miSVM           &78           &70.2       &82.2       &58.2       &78.4\\
%%MissSVM         &87.6           &80.0       &N/A        &N/A        &N/A\\
%EM-DD           &84.8           &84.9       &78.3       &56.1       &72.1\\
%PPMM            &$\bm{95.6}$           &81.2       &82.4       &60.3       &82.4\\
%%\hline
%\bottomrule[1.5pt]
%\end{tabular}
%\end{center}
%\end{table}

The widely-used MUSK datasets described in~\cite{DietterichLL97} for drug prediction are the benchmarks in nearly every previous MIL algorithm. Both of the datasets, MUSK1 and MUSK2, are composed of representations of molecules (bags) in multiple low-energy conformations (instances). Each conformation is described by a 166-dimensional feature vector derived from its surface properties. MUSK1 contains 476 instances divided into 47 positive bags and 45 negative bags, while MUSK2 owns approximately 6600 instances grouped into 39 positive bags and 63 negative bags. Another difference of these two datasets is that MUSK2 consists of more fraction of negative instances in a bag.

For this task, we set $\lambda=0.05, \beta=1.5$ and $m_0=0.5$ in the proposed algorithm. \textit{For all our experiments including this and the following tasks, we fix the $p_0$ in Eq.~\eqref{eq:Lins} to $0.5$ and the maximum iteration $T$ to 2000 by default if we don't particularly point out.} We compare our results with miSVM and MISVM proposed in~\cite{andrews2002support} in Table~\ref{table:MUSK}, which show that both MISVM and RMI-SVM achieve a similar accuracy on MUSK1 dataset and outperform miSVM by a few percent. Furthermore on MUSK2 dataset, RMI-SVM performs marginally better than miSVM, which is susceptible to local minima. Note that the results of miSVM and MISVM are implemented via linear kernel for fair comparison with RMI-SVM.

\begin{table}[ht]
\caption{Average prediction accuracy (\%) via ten times 10-fold cross validation on MUSK datasets. Please note that we all adopt linear kernels for fair comparison.}\label{table:MUSK}
% The standard deviations of MISVM and miSVM are not available.
\begin{center}
\begin{tabular}{lccc}\toprule
Dataset     &MISVM      &miSVM      &RMI-SVM\\
\hline
MUSK1       &80.4       &78.0       &$\bm{80.8}$\\
MUSK2       &77.5         &70.2        &$\bm{82.4}$\\
\bottomrule
\end{tabular}
\end{center}
\vspace{-0.5cm}
\end{table}

\subsection{Automatic Image Annotation}\label{sec:Annotation}

Widely applied to image retrieval systems, this task is the process by which an intelligent system automatically assigns context information in the form of \emph{keywords} to digital images. An image (bag) contains a set of regions/segments (instances) which denote different visual objects. Assuming that a user is searching for a target object, an image is regarded as a relevant retrieval if only one of its regions is relevant, while other regions are relevant or not.

We perform three classification experiments on ``elephant'', ``fox'' and ``tiger'' classes in the Corel dataset \cite{carson1999blobworld}. More specifically, each image (bag) consists of plenty of segments (instances) and a 320-dimensional feature is extracted to represent the color, texture and shape characteristics of a segment. There are 100 positive/relevant images and 100 negative/irrelevant ones for each dataset. As for each image, the number of positive segments (instances) is approximately the same with that of negative ones.

In this task, all instance feature are preprocessed by $L2$ normalization as input. The parameters are given as $\lambda=0.02, \beta=5$ and $m_0=2$. We compare our method with miGraph and MIGraph in~\cite{migraph}, miFV in~\cite{miFV}, miSVM and MISVM in~\cite{DietterichLL97}, EM-DD in~\cite{EMDD}, MILES~\cite{miles}, MIForests in~\cite{MIForests} and PPMM in~\cite{PPMM} via ten times 10-fold cross validation and report the average results and corresponding standard deviation in Table~\ref{table:Corel}. The results of MI-Kernel was taken from~\cite{migraph}. Note that some standard deviations in former studies are not available. RMI-SVM achieves the best results on the three datasets.

\begin{table}[ht]
\caption{Average prediction accuracy (\%) via ten times 10-fold cross validation on benchmarks. Some standard deviations in former approaches are not available.}\label{table:Corel}
\begin{center}
\begin{tabular}{lccc}\toprule
%\hline
Algorithm       &Elephant   &Fox        &Tiger\\
\hline
RMI-SVM         &$\bm{87.8}{\bm{\pm0.7}}$       &$\bm{63.6}$$\bm{\pm 2.8}$       &$\bm{87.9}$$\bm{\pm 0.9}$\\
%\hline
MIGraph         &85.1$\pm2.8$       &61.2$\pm 1.7$      &81.9$\pm 1.5$\\
%\hline
miGraph         &86.8$\pm0.7$       &61.6$\pm 2.8$      &86.0$\pm 1.6$\\
%\hline
miFV            &85.2$\pm{0.8}$     &62.1$\pm 1.1$      &81.3$\pm 0.8$\\
MI-Kernel       &84.3$\pm1.6$       &60.3$\pm 1.9$      &84.2$\pm 1.0$\\
%\hline
MISVM           &81.4       &57.8       &84.0\\
miSVM           &82.2       &58.2       &78.4\\
EM-DD           &78.3       &56.1       &72.1\\
PPMM            &82.4       &60.3       &82.4\\
{MIForests}\footnotemark[1]       &84         &{$\bm{64}$}         &{82}\\
{MILES}\footnotemark[1]       &81         &{62}         &{80}\\
%\hline
\bottomrule
\end{tabular}
\end{center}
\vspace{-0.5cm}
\end{table}

\footnotetext[1]{The results are reported in integer over 5 runs in~\cite{MIForests}.}

\subsection{Text Categorization}\label{sec:Text}
The task is to assign predefined categories to text documents. A document (bag) may be labeled as relevant to certain topic only if some unspecified paragraphs/keywords(instances) of it are relevant. In other words, a document is usually regarded as irrelevant if there are no relevant paragraphs/keywords. Therefore, document classification can be naturally formulated as a multiple instance problem.

We test the proposed method on datasets from text categorization. The evaluated datasets are randomly split and subsampled from the original TREC9 dataset. Compared with those datasets used in Sec.~\ref{sec:Drug} and~\ref{sec:Annotation}, the representation is extremely sparse and high-dimensional with more than 66000 dimension but less than 32 non-zero values, which makes them challenging datasets.

In this task, we set parameters as $\lambda=0.0003, \beta=4$ and $m_0=2$ and all data is $L2$ normalized. During the experiments, we find that slight changes in the parameters make minor difference to the final average accuracy. Results of the proposed approach are reported in Table~\ref{table:TREC9}. We achieve the best results over the previous methods in all the seven subsets. The average classification accuracy is improved by more than 3 percent. Note that RMI-SVM with linear kernel consistently performs better than both miSVM and MISVM whatever the kernels they adopt.

\begin{table*}[htp]\label{Tab Test}
\small
\caption{Classification accuracy (\%) of methods on seven subsets from TREC9. The standard deviations of other methods are not available.}\label{table:TREC9}
\vspace{-0.4cm}
\begin{center}
\begin{tabular}{cccccccccc}\toprule% l r c the aligned format
%\hline
Dataset &Dims       &EM-DD  &\multicolumn{3}{c}{miSVM}     &\multicolumn{3}{c}{MISVM} &RMI-SVM\\\cline{4-6}\cline{7-9}
Category&Ins/Feat   &       &linear & poly  & rbf           &linear & poly  & rbf       &\\ \midrule
TST1    &3224/66552 &85.8   &93.6   &92.5   &90.4           &93.9   &93.8   &93.7       &$\bm{95.0\pm{1.0}}$\\
TST2    &3344/66153 &84.0   &78.2   &75.9   &74.3           &84.5   &84.4   &76.4       &$\bm{86.3\pm{0.8}}$\\
TST3    &3246/66144 &69.0   &87.0   &83.3   &69.0           &82.2   &85.1   &77.4       &$\bm{87.9\pm{0.6}}$\\
TST4    &3391/68085 &80.5   &82.8   &80.0   &69.6           &82.4   &82.9   &77.3       &$\bm{85.3\pm{1.0}}$\\
TST7    &3367/66823 &75.4   &81.3   &78.7   &81.3           &78.0   &78.7   &64.5       &$\bm{82.3\pm{0.8}}$\\
TST9    &3300/66627 &65.5   &67.5   &65.6   &55.2           &60.2   &63.7   &57.0       &$\bm{71.2\pm{0.7}}$\\
TST10   &3453/66082 &78.5   &79.6   &78.3   &52.6           &79.5   &81.0   &69.1       &$\bm{83.9\pm{0.8}}$\\
\hline
Average &3332/66638 &77.0   &81.4   &79.2   &70.3           &80.1   &81.4   &73.6       &$\bm{84.8\pm{0.8}}$\\
\bottomrule
\end{tabular}
\end{center}
\vspace{-0.5cm}
\end{table*}

\begin{figure}
\vspace{-0.25cm}
  \centering
  \includegraphics[width=1\linewidth]{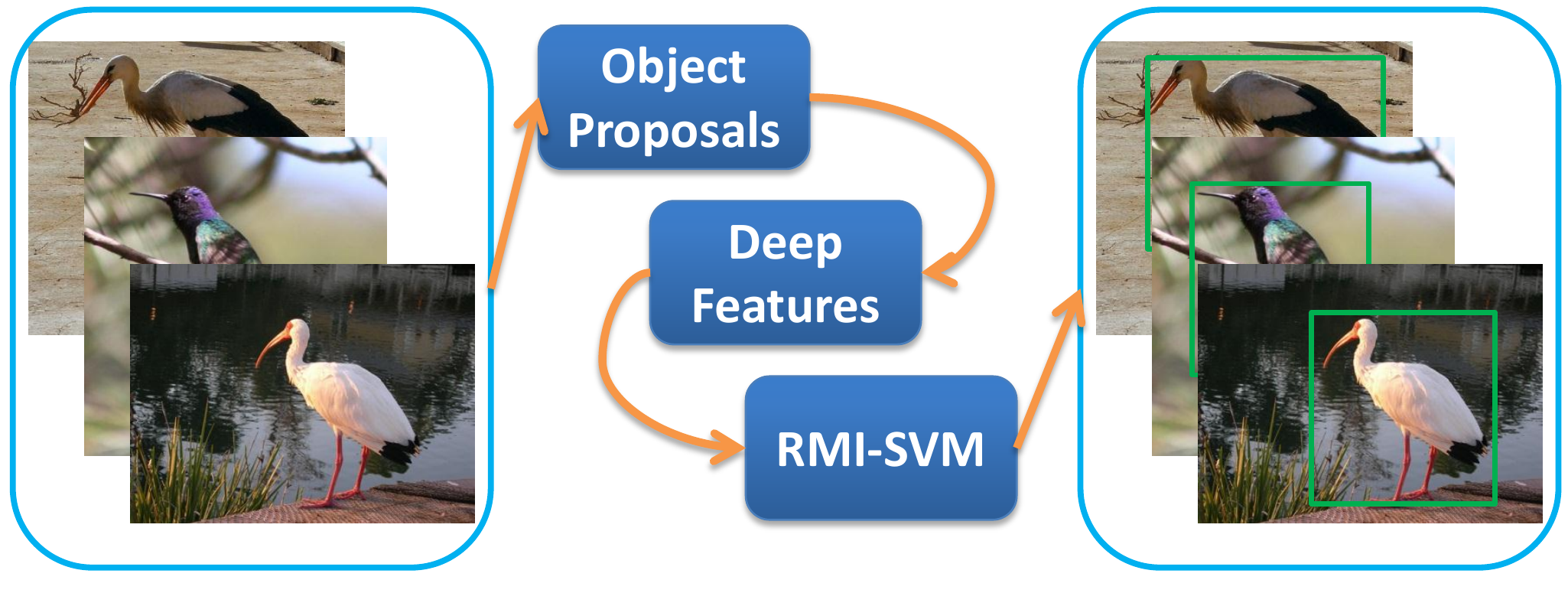}\\
  \caption{Illustration of our object discovery pipeline in the experiments. At first, the object proposal method Edgebox extracts candidate object regions. Then, for every candidate, its DCNN feature is extracted. At last, RMI-SVM identifies positive instances as object discovery results.}\label{fig:pipeline}
\vspace{-0.5cm}
\end{figure}

\section{Experiments of Object Discovery in the Wild}\label{object discovery}
\vspace{-0.2cm}
\subsection{Datasets and Evaluation Criteria}

%\vspace{-0.1cm}

In this section, we perform weakly-supervised object discovery in natural images following the pipeline shown in Fig.~\ref{fig:pipeline}.
% \textcolor{red}{The NMS threshold of Edgebox is set to 0.55}.
Given a set of images, we firstly utilize Edgebox~\cite{Edgebox} to capture plenty of windows/patches as object proposals. This strategy turns the object discovery problem into a well-defined MIL problem, in which an image is a bag, an object proposal is an instance, and image label is used as bag label. Then, a pre-trained DCNN is applied to extract the rich semantic feature for each object proposal. Here, we use the BVLC AlexNet model provide in Caffe Model Zoo \cite{jia2014caffe}. Furthermore, we treat the images containing a shared object as the positive set and randomly select images from the remaining images as negative. At last, after that the models adopting the proposed method are learnt, we report the object proposals with maximal value predicted by RMI-SVM as the detected object. The final results evaluated via CorLoc measure \cite{LearnwithGeneric}, which is the percentage of the correct location of objects under the Pascal criteria (intersection over union (IoU) $>0.5$ between detected bounding boxes and the ground truth).

The popular Pascal 2006 and 2007 datasets \cite{Pascal07} are extremely challenging and have been widely used as the benchmarks to evaluate object discovery methods. Following the protocol of~\cite{LearnwithGeneric}, two subsets are taken from Pascal 2006 and 2007 \emph{train+val} dataset, which are then divided into various of class and view combinations. The two subsets are referred as \textit{Pascal06-all} and \textit{Pascal07-all} below, respectively. There are in total 2047 images divided into 45 class/viewpooint combinations in Pascal07-all while total 2184 images from 33 class/viewpoint in Pascal06-all. Besides of Pascal06-all and Pascal07-all, recent methods start to focus on the 20 classes Pascal 2007 training set (denoted as Pascal 2007) without considering view variations, which makes the object discovery task more challenging. Thus, in the experiments, we have three different testing sets: Pascal06-all, Pascal07-all, and Pascal 2007. Following the common setting~\cite{LearnwithGeneric}, for the three sets, we use all images that contain at least one object instance not marked as truncated or difficult in the ground truth.

We utilize the Structured Edge Detection Toolbox in~\cite{Edgebox} to extract a large number of object proposals. The parameters are given via the step size of 0.65, Non-maximal suppression (NMS) threshold of 0.55, minimum score of boxes to detect of 0.1 and maximal number of boxes to detect of 400. {For two object proposals in NMS, if the ratio of intersect area to union area is greater than a given threshold, then the proposal with the lower score is suppressed.} As for the DCNN feature extraction, we adopt the exact output of the $fc6$ layer, whose dimension is 4096. On Pascal07-all dataset, we set $\lambda=0.0015, \beta=5$ and $m_0=1.2$, while $\lambda=0.0015, \beta=6$ and $m_0=0.2$ on Pascal06-all dataset.

\begin{figure*}
  \centering
  \includegraphics[width=0.9\textwidth]{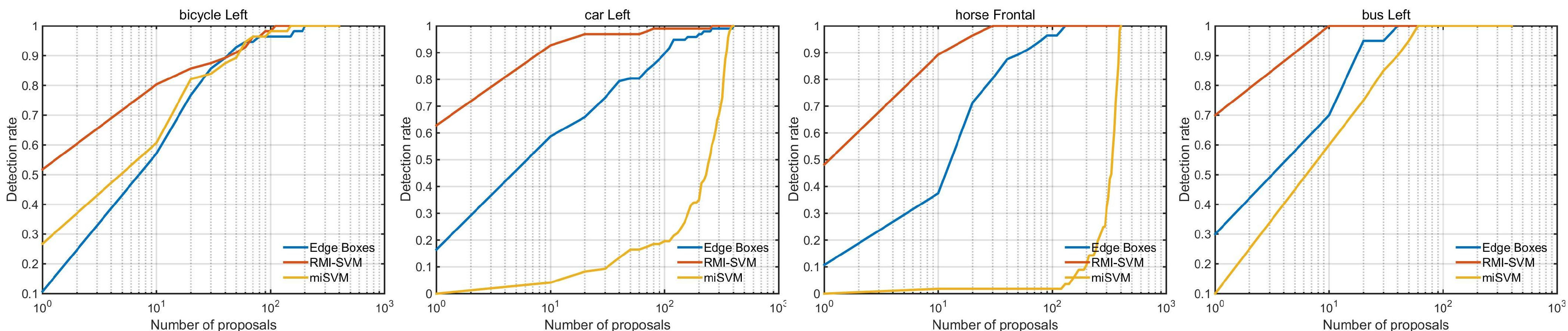}\\
  \caption{Detection rates when changing the number of detections/proposals on four class/viewpoint combinations. These combinations are, from left to right and top to bottom, Bicycle/Left, car/Left, House/Frontal, Bus/Left.}\label{fig:detection_rate}
\end{figure*}

\subsection{Comparison to State-of-the-arts}
\subsubsection{Pascal06-all and Pascal07-all}

\begin{figure*}
  \centering
  \includegraphics[width=0.78\textwidth]{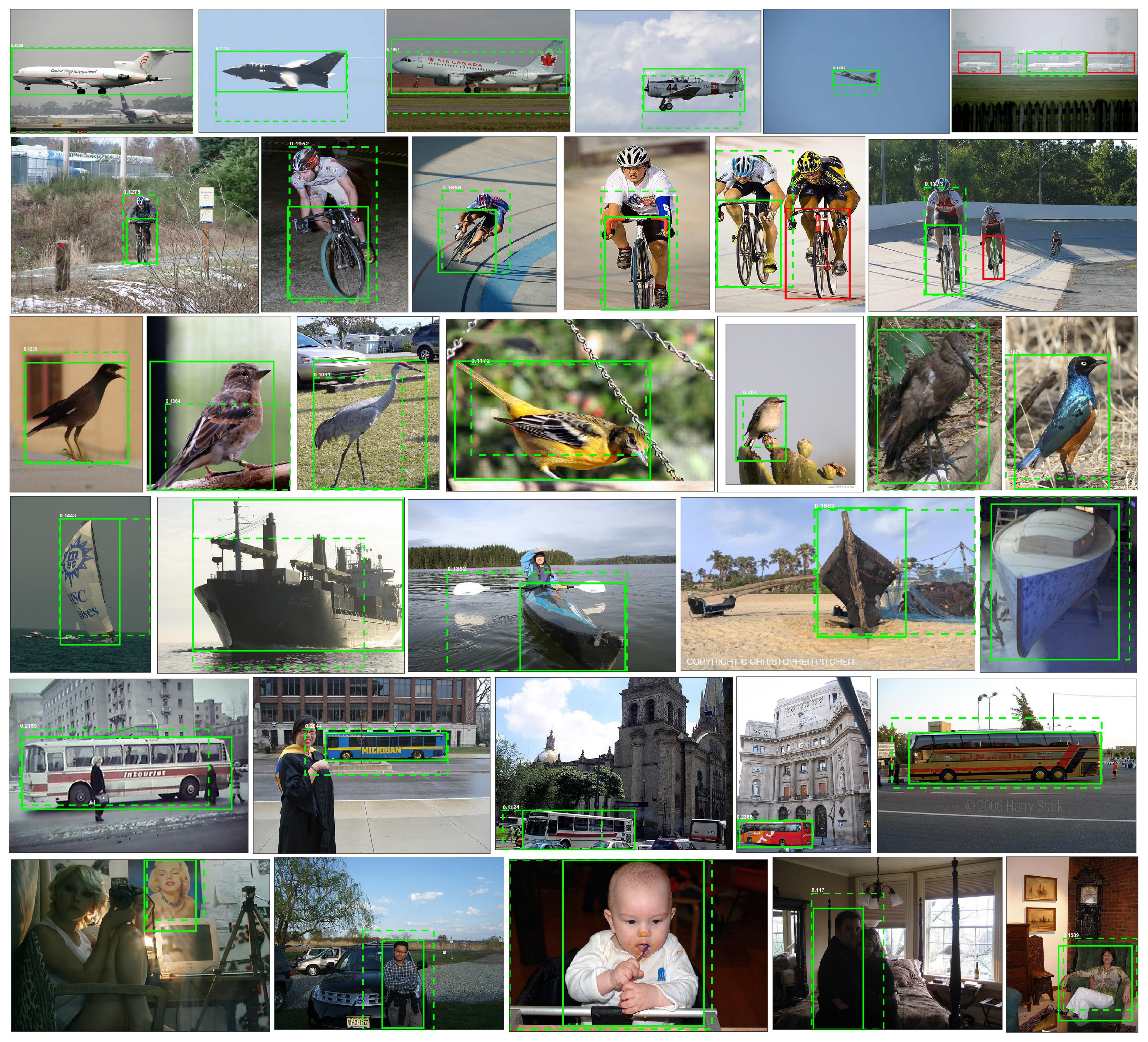}\\
  \caption{Results of object discovery on several class/viewpoint combinations on Pascal07-all set. Each row denotes one combination. These combinations are, from top row to bottom, Aeroplane/Left, Bicycle/Frontal, Bird/Right, Boat/Frontal, Bus/Left, Person/Frontal. It is worth noting that the solid green rectangle denotes the matched ground truth; the dashed green rectangle denotes the matched detection; and the solid red rectangle denotes the missed ground truth. Best viewed in color.}\label{fig:detection_result}
\end{figure*}

The results of the proposed method on Pascal06-all and Pascal07-all are compared with the former state-of-the-art works and shown in Table~\ref{table:pascal0607}. Our method consistently yields better performance than other former state-of-the-art approaches on the two datasets. The CorLoc measures have been improved by 4\% and 9\% on Pascal06-all and Pascal07-all respectively. Some object discovery results are shown in Fig.~\ref{fig:detection_result}.

The CorLoc measure is not accurate enough since there may have more than one object of interests in image. To better characterize the discovery performance, we plot the detection v.s. the number of detections curve in Fig.~\ref{fig:detection_rate}, and compare our method to miSVM and Edgebox. In the compared classes, our RMI-SVM can consistently and significantly improves Edgebox; but miSVM failed. The curves also show that our RMI-SVM is more robust than miSVM.

\begin{table}
\small
\caption{Object discovery results evaluated via CorLoc on Pascal06-all and Pascal07-all.}\label{table:pascal0607}
\vspace{-0.6cm}
\begin{center}
\begin{tabular}{p{1.6cm}<{\centering}p{0.4cm}<{\centering}p{0.5cm}<{\centering}p{0.7cm}
<{\centering}p{0.9cm}<{\centering}p{0.9cm}<{\centering}p{0.9cm}<{\centering}}\toprule
Dataset&Ours&bMCL&ADMM&MIForests&WSDPM&Deselaers\\
                &       &\cite{bMCL}   &\cite{RobustSubspace}  &\cite{MIForests}    &\cite{WSDPM}   &\emph{et.al.}\cite{LearnwithGeneric}\\
\hline
Pascal06-all&$\bm{53}$&45&43&36&N/A&49\\
Pascal07-all&$\bm{37}$&31&27&25&30&28\\
\bottomrule
\end{tabular}
\end{center}
\vspace{-0.5cm}
\end{table}

\begin{table*}[htp]
\centering
\caption{Object discovery results evaluated via CorLoc of all 20 classes on Pascal 2007 training set. Note that the last column is the average CorLoc of all 20 classes. The best result of each class is emphasized in bold.}\label{table:pascal0720}
\vspace{-0.3cm}
\begin{center}
\resizebox{2.1\columnwidth}{!}{
\begin{tabular}{p{2.7cm}<{\centering}p{0.4cm}<{\centering}p{0.4cm}p{0.4cm}<{\centering}p{0.4cm}<{\centering}p{0.4cm}
<{\centering}p{0.4cm}<{\centering}p{0.4cm}<{\centering}p{0.4cm}<{\centering}p{0.4cm}<{\centering}p{0.4cm}<{\centering}p{0.4cm}<{\centering}
p{0.4cm}<{\centering}p{0.4cm}<{\centering}p{0.4cm}<{\centering}p{0.4cm}<{\centering}p{0.4cm}<{\centering}p{0.4cm}<{\centering}
p{0.4cm}<{\centering}p{0.4cm}<{\centering}p{0.4cm}<{\centering}p{0.4cm}<{\centering}}\toprule
Algorithm     &aero   &bicy &bird &boa &bot &bus &car &cat &cha &cow &dtab &dog &hors &mbik &pers &plnt &she &sofa &trai &tv &Av.\\
\hline
Multi-fold MIL\cite{MfMIL} &56.6 &58.3 &28.4 &$\bm{20.7}$ &$\bm{6.8}$ &$\bm{54.9}$ &69.1 &20.8 &$\bm{9.2}$ &50.5 &10.2 &29.0 &58.0 &64.9 &$\bm{36.7}$ &18.7 &$\bm{56.5}$ &13.2 &54.9 &$\bm{59.4}$ &38.8\\
Shi {\emph{et al.}}'13\cite{ShiHX13} &$\bm{67.3}$ &54.4 &34.3 &17.8 &1.3 &46.6 &60.7 &$\bm{68.9}$ &2.5 &32.4 &16.2 &$\bm{58.9}$ &51.5 &64.6 &18.2 &3.1 &20.9 &34.7 &$\bm{63.4}$ &5.9 &36.2\\
Siva {\emph{et al.}}'13\cite{SivaCVPR13} &-- &-&-&-&-&-&-&-&-&-&-&-&-&-&-&-&-&-&-&-&32.0\\
Siva\&Xiang'11\cite{SivaICCV11} &42.4 &46.5 &18.2 &8.8 &2.9 &40.9 &$\bm{73.2}$ &44.8 &5.4 &30.5 &19.0 &34.0 &48.8 &$\bm{65.3}$ &8.2 &9.4 &16.7 &32.3 &54.8 &5.5 &30.4\\
Siva {\emph{et al.}}'12~\cite{SivaRX12}  &45.8 &21.8 &30.9 &20.4 &5.3 &37.6 &40.8 &51.6 &7.0 &29.8 &27.5 &41.3 &41.8 &47.3 &24.1 &12.2 &28.1 &32.8 &48.7 &9.4 &30.2\\
\hline
Ours&37.7   &$\bm{58.8}$ &$\bm{39.0}$ &4.7 &4.0 &48.4 &70.0 &63.7 &9.0 &$\bm{54.2}$ &$\bm{33.3}$ &37.4 &$\bm{61.6}$ &57.6 &30.1 &$\bm{31.7}$ &32.4 &$\bm{52.8}$ &49.0 &27.8 &$\bm{40.2}$\\
\bottomrule
\end{tabular}
}
\end{center}
\end{table*}
\vspace{-0.5cm}

\subsubsection{Pascal 2007}
\vspace{-0.3cm}

The object discovery results on the 20 classes Pascal 2007 set measure by CorLoc are given in Table~\ref{table:pascal0720}. Recent weakly-supervised detectors are compared, including the previous state-of-the-art method named Multi-fold MIL \cite{MfMIL}. The average CorLoc of Multi-fold MIL is 38.8\%. It uses the advanced fisher vector coding \cite{perronnin2010improving} to extract object feature. Our RMI-SVM based method improves the average CorLoc to 40.2\% and wins in 7 out of 20 classes. The good results indicate that: (1) The proposed RMI-SVM is more robust and effective than other MIL algorithms, such as, the Multi-fold MIL and miSVM; (2) The DCNN feature used in our paper is very robust to view variation, since DCNN is learnt from the huge ImageNet dataset.
\vspace{-0.4cm}

\subsection{Improvement of detection performance over Edgebox}
\vspace{-0.3cm}
Edgebox is a method for generating object bounding box proposals using informative edges. However, it is imperative to extract a large number of proposals to reach a high detection rate. Given the labels of each image, we conduct experiments to demonstrate that RMI-SVM can assist Edgebox on detection task to a great margin even with a few number of object proposals. Under the criteria of IoU $>0.5$, we show the detection rates on Pascal07-all when varying the number of detections. Several results of different class/viewpoint combinations are given in Fig.~\ref{fig:detection_rate}. We can find that the detection rates are greatly improved via using the weakly supervised information.

\vspace{-0.2cm}
\subsection{Comparison to miSVM}
\vspace{-0.2cm}
We compare the effectiveness of the proposed RMI-SVM with the conventional miSVM on the object discovery task.
As stated in~\ref{object discovery}, we first make use of the Edgebox to extract object proposals. Then deep representation is captured using Convolutional Neural Network, which is finally the process of multiple instance learning. To keep fair comparison, we replace the RMI-SVM with miSVM to guarantee the exactly same features as input in the final learning step. Results are given in Table~\ref{table:miSVMObject}, which demonstrates that the learning ability of RMI-SVM is superior over miSVM to a great margin under the MIL constraints. The Liblinear~\cite{liblinear} toolbox is chosen in the implementation of miSVM.

As shown in Fig.~\ref{fig:detection_rate}, the RMI-SVM is superior to miSVM when choosing the number of proposals as 1, which is exactly the CorLoc evaluation. It obviously shows that miSVM fails in learning the common attributes in the same class/viewpoint. In miSVM framework, all the instances in positive bag are initialized as positive, followed by updating instance labels in each iteration. This learning strategy seems reasonable in Sections~\ref{sec:Drug},~\ref{sec:Annotation} and~\ref{sec:Text} since the ratio of positive instances to negative ones is approximately 1, except the MUSK2 dataset where the ratio is $\frac{1}{6}$. When the number of positive instances makes up to quite a large portion in a positive bag, miSVM can find a hyperplane that divides the negative instances as true negative. However, positive proposals in positive images hold a small portion after the NMS on object discovery, usually less than $\frac{1}{20}$. Even though miSVM wrongly classifies the negative instances in negative bag as positive, it considers little penalty in each iteration. Thus, miSVM which is always susceptible to local minima cannot distinguish background windows/patches well from the shared object proposals. However, RMI-SVM accounts for the penalty of false positive via the term in Eq.~\eqref{eq:Lbagpart1}. Thus, RMI-SVM can well separate the background proposals from the common object proposals.

\begin{table}[ht]
\small
\caption{Comparison between RMI-SVM and miSVM via CorLoc evaluation and running time on object discovery experiments.}\label{table:miSVMObject}
\vspace{-0.3cm}
\begin{center}
\begin{tabular}{lcc}\toprule
Evaluation  &RMI-SVM   &miSVM\\
\hline
CorLoc (\%) on Pascal06-all    &$\bm{53}$     &30\\
CorLoc (\%) on Pascal07-all    &$\bm{37}$     &20\\
Running time (s) on Pascal07-all    &$\bm{854}$     &4300\\
\bottomrule
\end{tabular}
\end{center}
\vspace{-0.8cm}
\end{table}

Furthermore, we experimentally compare the time complexity of RMI-SVM with miSVM. On Pascal07-all dataset for object discovery, it takes RMI-SVM around 854 seconds to learn 45 models for all combinations, while miSVM spends more than 4300 seconds. RMI-SVM is 5 times efficient than the conventional miSVM. The performance gain in running time should be owned to our novel formulation and the fast SGD. The SGD in RMI-SVM randomly uses only one bag. As for in every iteration of miSVM, it takes all instances of all bags as input, which is the crucial issue of time consuming.
Other EM-style MIL methods, \emph{e.g.}, MILBoost, have the same mechanism, and are less efficient than our RMI-SVM.

%
%\subsection{Feature layer selection on object discovery}
%A Convolutional Neural Network (CNN) is comprised of several convolution layers (often with a subsampling step) and then followed by one or more fully connected layers. In the feature extraction step, we use a CNN structure of 5 convolutional and 3 fully connected layers. It is interesting to explore the representation capabilities and effectiveness using output of different layers as features. The last 3 fully connected layers are referred as fc6, fc7 and fc8 below respectively, where fc8 denotes the last layer. Usually the output of the whole network owns high-level context information while the convolution layers hold powerful generalization. Results are displayed in Table~\ref{table:layerselection}. As we can see, the fc6 that connects the last convolutional layer is much preferable as input features.
%
%\begin{table}[ht]
%\caption{CorLoc results of different layers as features in RMI-SVM on object discovery.}\label{table:layerselection}
%\begin{center}
%\begin{tabular}{lccc}\toprule
%Dataset         &fc6            &fc7        &fc8\\
%\hline
%Pascal06-all    &$\bm{53}$      &-         &-\\
%Pascal07-all    &$\bm{35}$      &25         &15\\
%\bottomrule
%\end{tabular}
%\end{center}
%\end{table}

\vspace{-0.30cm}
\section{Conclusion}
\vspace{-0.3cm}
In this paper, we have proposed a novel formulation for MIL and applied it for robust weakly-supervised object discovery. Different from the traditional EM-style MIL solutions, we relax the highly combinatorial MIL optimization problem into a convex program and solve it efficiently using SGD.
Our idea of solving MIL in a relaxed formulation is general. More complex discriminative model and model regularization method, \emph{e.g.}, deep neural networks, can be adopted. Besides of object discovery, RMI-SVM can also be used to solve other recognition tasks, such as visual tracking, image classification, and learning part-based object detection model.
\vspace{-0.30cm}
\section{Acknowledgement}
\vspace{-0.3cm}
This work was supported by National Natural Science Foundation of China (NSFC) (NO. 61503145, NO. 61222308 and NO. 61573160), the Fundamental Research Funds for the Central Universities (HUST 0118181099), and CCF-Tencent RAGR (NO. 20140116).

\begin{footnotesize}
\bibliographystyle{abbrv}
\bibliography{egbib}
\end{footnotesize}

\end{document}